# Comlex Syntax: Building a Computational Lexicon

Ralph Grishman, Catherine Macleod, and Adam Meyers

Computer Science Department, New York University
715 Broadway, 7th Floor, New York, NY 10003, U.S.A.
{grishman,macleod,meyers}@cs.nyu.edu

## Abstract

We describe the design of Comlex Syntax, a computational lexicon providing detailed syntactic information for approximately 38,000 English headwords. We consider the types of errors which arise in creating such a lexicon, and how such errors can be measured and controlled.

## 1 Goal

The goal of the Comlex Syntax project is to create a moderately-broad-coverage lexicon recording the syntactic features of English words for purposes of computational language analysis. This dictionary is being developed at New York University and is to be distributed by the Linguistic Data Consortium, to be freely usable for both research and commercial purposes by members of the Consortium.

In order to meet the needs of a wide range of analyzers, we have included a rich set of syntactic features and have aimed to characterize these features in a relatively theory-neutral way. In particular, the feature set is more detailed than those of the major commercial dictionaries, such as the Oxford Advanced Learner's Dictionary (OALD) [4] and the Longman Dictionary of Contemporary English (LDOCE) [8], which have been widely used as a source of lexical information in language analyzers.[1] In addition, we have aimed to be more comprehensive in capturing features (in particular, subcategorization features) than commercial dictionaries.

## 2 Structure

The word list was derived from the file prepared by Prof. Roger Mitton from the Oxford Advanced Learner's Dictionary, and contains about 38,000 head forms, although some purely British terms have been omitted. Each entry is organized as a nested set of typed feature-value lists. We currently use a Lisp-like parenthesized list notation, although the lexicon could be readily mapped into other forms, such as SGML-marked text, if desired.

Some sample dictionary entries are shown in Figure 1. The first symbol gives the part of speech; a word with several parts of speech will have several dictionary entries, one for each part of speech. Each entry has an :orth feature, giving the base form of the word. Nouns, verbs, and adjectives with irregular morphology will have features for the irregular forms :plural, :past, :past-part, etc. Words which take complements will have a subcategorization (:subc) feature. For example, the verb "abandon" can occur with a noun phrase followed by a prepositional phrase with the preposition "to" (e.g., "I abandoned him to the linguists.") or with just a noun phrase complement ("I abandoned the ship."). Other syntactic features are recorded under :features. For example, the noun "abandon" is marked as (countable :pval ("with")), indicating that it must appear in the singular with a determiner unless it is preceded by the preposition "with".

### 2.1 Subcategorization

We have paid particular attention to providing detailed subcategorization information (information about complement structure), both for verbs and for those nouns and adjectives which do take complements. In order to insure the completeness of our codes, we studied the coding employed by several other major lexicons, including the Brandeis Verb Lexicon[2], the ACQUILEX Project [10], the NYU Linguistic String Project [9], the OALD, and LDOCE, and, whenever feasible, have sought to incorporate distinctions made in any of these dictionaries. Our resulting feature system includes 92 subcategorization features for verbs, 14 for adjectives, and 9 for nouns. These features record differences in grammatical functional structure as well as constituent structure. In particular, they capture four different types of control: subject control, object control, variable control, and arbitrary control. Furthermore, the notation allows us to indicate that a verb may have different control features for different complement structures, or even for different prepositions within the complement. We record, for example, that "blame ... on" involves arbitrary control ("He

---

[1] To facilitate the transition to COMLEX by current users of these dictionaries, we have prepared mappings from COMLEX classes to those of several other dictionaries.

[2] Developed by J. Grimshaw and R. Jackendoff.

```
(verb         :orth "abandon"   :subc ((np-pp :pval ("to")) (np)))
(noun         :orth "abandon"   :features ((countable :pval ("with"))))
(prep         :orth "above")
(adverb       :orth "above")
(adjective    :orth "above"     :features ((ainrn) (apreq)))
(verb         :orth "abstain"   :subc ((intrans)
                                       (pp :pval ("from"))
                                       (p-ing-sc :pval ("from"))))
(verb         :orth "accept"    :subc ((np) (that-s) (np-as-np)))
(noun         :orth "acceptance")
```

Figure 1: Sample COMLEX Syntax dictionary entries.

blamed the country's health problems on eating too much chocolate."), whereas "blame for" involves object control ("He blamed John for going too fast.").

The names for the different complement types are based on the conventions used in the Brandeis verb lexicon, where each complement is designated by the names of its constituents, together with a few tags to indicate things such as control phenomena. Each complement type is formally defined by a frame (see Figure 2). The frame includes the constituent structure, :cs, the grammatical structure, :gs, one or more :features, and one or more examples, :ex. The constituent structure lists the constituents in sequence; the grammatical structure indicates the functional role played by each constituent. The elements of the constituent structure are indexed, and these indices are referenced in the grammatical structure field (in vp-frames, the index "1" in the grammatical structures always refers to the surface subject of the verb).

Three verb frames are shown in Figure 2. The first, s, is for full sentential complements with an optional "that" complementizer. The second and third frames both represent infinitival complements, and differ only in their functional structure. The to-inf-sc frame is for subject-control verbs — verbs for which the surface subject is the functional subject of both the matrix and embedded clauses. The notation :subject 1 in the :cs field indicates that the surface subject is the subject of the embedded clause, while the :subject 1 in the :gs field indicates that it is the subject of the matrix clause. The indication :features (:control subject) provides this information redundantly; we include both indications in case one is more convenient for particular dictionary users. The to-inf-rs frame is for raising-to-subject verbs — verbs for which the surface subject is the functional subject only of the embedded clause. The functional subject position in the matrix clause is unfilled, as indicated by the notation :gs (:subject () :comp 2).

## 3 Methods

Our basic approach has been to create an initial lexicon manually and then to use a variety of resources, both commercial and corpus-derived, to refine this lexicon. Although methods have been developed over the last few years for automatically identifying some subcategorization constraints through corpus analysis [2,5], these methods are still limited in the range of distinctions they can identify and their ability to deal with low-frequency words. Consequently we have chosen to use manual entry for creation of our initial dictionary.

The entry of lexical information is being performed by four graduate linguistics students, referred to as elves ("elf" = enterer of lexical features). The elves are provided with a menu-based interface coded in Common Lisp using the Garnet GUI package, and running on Sun workstations. This interface also provides access to a large text corpus; as a word is being entered, instances of the word can be viewed in one of the windows. Elves rely on citations from the corpus, definitions and citations from any of several printed dictionaries and their own linguistic intuitions in assigning features to words.

Dictionary entry began in April 1993. An initial dictionary containing entries for all the nouns, verbs and adjectives in the OALD was completed in May, 1994.[3]

We expect to check this dictionary against several sources. We intend to compare the manual subcategorizations for verbs against those in the OALD, and would be pleased to make comparisons against other broad-coverage dictionaries if those can be made available for this purpose. We also intend to make comparisons against several corpus-derived lists: at the very least, with verb/preposition and verb/particle pairs with high mutual information [3] and, if possible, with the results of recently-developed procedures for extracting subcategorization frames from corpora [2,5]. While this corpus-derived information may not be detailed or accurate enough for fully-automated lexicon

---

[3]No features are being assigned to adverbs in the initial lexicon

```
(vp-frame s          :cs ((s 2 :that-comp optional))
                     :gs (:subject 1 :comp 2)
                     :ex "they thought (that) he was always late")

(vp-frame to-inf-sc  :cs ((vp 2 :mood to-infinitive :subject 1))
                     :features (:control subject)
                     :gs (:subject 1 :comp 2)
                     :ex "I wanted to come.")

(vp-frame to-inf-rs  :cs ((vp 2 :mood to-infinitive :subject 1))
                     :features (:raising subject)
                     :gs (:subject () :comp 2)
                     :ex "they seemed to wilt.")
```

Figure 2: Sample COMLEX Syntax subcategorization frames.

creation, it should be most valuable as a basis for comparisons.

## 4 Types and Sources of Error

As part of the process of refining the dictionary and assuring its quality, we have spent considerable resources on reviewing dictionary entries and on occasion have had sections coded by two or even four of the elves. This process has allowed us to make some analysis of the sources and types of error in the lexicon, and how these errors might be reduced. We can divide the sources of error and inconsistency into four classes:

1. **errors of classification**: where an instance of a word is improperly analyzed, and in particular where the words following a verb are not properly identified with regard to complement type. Specific types of problems include misclassifying adjuncts as arguments (or vice versa) and identifying the wrong control features. Our primary defenses against such errors have been a steady refinement of the feature descriptions in our manual and regular group review sessions with all the elves. In particular, we have developed detailed criteria for making adjunct/argument distinctions [6].

   A preliminary study, conducted on examples (drawn at random from a corpus not used for our concordance) of verbs beginning with "j", indicated that elves were consistent 93% to 94% of the time in labeling argument/adjunct distinctions following our criteria and, in these cases, rarely disagreed on the subcategorization. In more than half of the cases where there was disagreement, the elves separately flagged these as difficult, ambiguous, or figurative uses of the verbs (and therefore would probably not use them as the basis for assigning lexical features). The agreement rate for examples which were not flagged was 96% to 98%.

2. **omitted features**: where an elf omits a feature because it is not suggested by an example in the concordance, a citation in the dictionary, or the elf's introspection. In order to get an estimate of the magnitude of this problem we decided to establish a measure of coverage or "recall" for the subcategorization features assigned by our elves. To do this, we tagged the first 150 "j" verbs from a randomly selected corpus from a part of the San Diego Mercury which was not included in our concordance and then compared the dictionary entries created by our lexicographers against the tagged corpus. The results of this comparison are shown in Figure 3.

   The "Complements only" is the percentage of instances in the corpus covered by the subcategorization tags assigned by the elves and does not include the identification of any prepositions or adverbs. The "Complements only" would correspond roughly to the type of information provided by OALD and LDOCE[4]. The "Complements + Prepositions/Particles" column includes all the features, that is it considers the correct identification of the complement plus the specific prepositions and adverbs required by certain complements. The two columns of figures under "Complements + Prepositions/Particles" show the results with and without the enumeration of directional prepositions.

   We have recently changed our approach to the classification of verbs (like "run", "send", "jog", "walk", "jump") which take a long list of directional prepositions, by providing our entering program with a P-DIR option on the preposition list. This option will automatically assign a list of directional prepositions to the verb and thus will save time and eliminate errors of missing prepositions. In some cases this approach will provide a preposition list that is a little rich for a given verb but we have decided to err on the side of a slight overgeneration rather than risk missing any prepositions which actually occur. As you can see, the removal of the P-DIRs from consideration improves the individual elf scores.

---

[4]LDOCE does provide some prepositions and particles.

| elf # | Complements only | Complements + Prepositions/Particles | |
|---|---|---|---|
| | | without P-DIR | using P-DIR |
| 1 | 96% | 89% | 90% |
| 2 | 82% | 63% | 79% |
| 3 | 95% | 83% | 92% |
| 4 | 87% | 69% | 81% |
| elf av | 90% | 76% | 84% |
| elf union | 100% | 93% | 94% |

Figure 3: Number of subcategorization features assigned to "j" verbs by different elves.

| elf # | Complements only | Complements + Prepositions/Particles | |
|---|---|---|---|
| | | without P-DIR | using P-DIR |
| 1 + 2 | 100% | 91% | 93% |
| 1 + 3 | 97% | 91% | 92% |
| 1 + 4 | 96% | 91% | 91% |
| 2 + 3 | 99% | 89% | 90% |
| 2 + 4 | 95% | 79% | 86% |
| 3 + 4 | 97% | 85% | 92% |
| 2-elf av | 97% | 88% | 91% |

Figure 4: Number of subcategorization features assigned to "j" verbs by pairs of elves.

The elf union score is the union of the lexical entries for all four elves. These are certainly numbers to be proud of, but realistically, having the verbs done four separate times is not practical. However, in our original proposal we stated that because of the complexity of the verb entries we would like to have them done twice. As can be seen in Figure 5, with two passes we succeed in raising individual percentages in all cases.

We would like to make clear that even in the two cases where our individual lexicographers miss 18% and 13% of the complements, there was only one instance in which this might have resulted in the inability to parse a sentence. This was a missing intransitive. Otherwise, the missed complements would have been analyzed as adjuncts since they were a combination of prepositional phrases and adverbials with one case of a subordinate conjunction "as".

We endeavored to make a comparison with LDOCE on the measurement. This was a bit difficult since LDOCE lacks some complements we have and combines others, not always consistently. For instance, our PP roughly corresponds to either L9 (our PP/ADVP) or prep/adv + T1 (e.g. "on" + T1) (our PP/PART-NP) but in some cases a preposition is mentioned but the verb is classified as intransitive. The straightforward comparison has LDOCE finding 73% of the tagged complements but a softer measure eliminating complements that LDOCE seems to be lacking (PART-NP-PP, P-POSSING, PP-PP) and allowing for a PP complement for "joke", although it is not specified, results in a percentage of 79.

We have adopted two lines of defense against the problem of omitted features. First, critical entries (particularly high frequency verbs) have been done independently by two or more elves. Second, we are developing a more balanced corpus for the elves to consult. Recent studies (e.g., [1]) confirm our observations that features such as subcategorization patterns may differ substantially between corpora. We began with a corpus from a single newspaper (San Jose Mercury News), but have since added the Brown corpus, several literary works from the Library of America, scientific abstracts from the U.S. Department of Energy, and an additional newspaper (the Wall Street Journal). In extending the corpus, we have limited ourselves to texts which would be readily available to members of the Linguistic Data Consortium.

3. **excess features**: when an elf assigns a spurious feature through incorrect extrapolation or analogy from available examples or introspection. Because of our desire to obtain relatively complete feature sets, even for infrequent verbs, we have permit-ted elves to extrapolate from the citations found. Such a process is bound to be less certain than the assignment of features from extant examples. However, this problem does not appear to be very severe. A review of the "j" verb entries produced by all four elves indicates that the fraction of spurious entries ranges from 2% to 6%.

4. **fuzzy features**: feature assignment is defined in terms of the acceptability of words in particular syntactic frames. Acceptability, however, is often not absolute but a matter of degree. A verb may occur primarily with particular complements, but will be "acceptable" with others.

    This problem is compounded by words which take on particular features only in special contexts. Thus, we don't ordinarily think of "dead" as being gradable (*"Fred is more dead than Mary."), but we do say "deader than a door nail". It is also compounded by our decision not to make sense distinctions initially. For example, many words which are countable (require a determiner before the singular form) also have a generic sense in which the determiner is not required (*"Fred bought apple." but "Apple is a wonderful flavor."). For each such problematic feature we have prepared guidelines for the elves, but these still require considerable discretion on their part.

These problems have emphasized for us the importance of developing a tagged corpus in conjunction with the dictionary, so that frequency of occurrence of a feature (and frequency by text type) will be available. We have done some preliminary tagging in parallel with the completion of our initial dictionary. We expect to start tagging in earnest in early summmer. Our plan is to begin by tagging verbs in the Brown corpus, in order to be able to correlate our tagging with the word sense tagging being done by the WordNet group on the same corpus [7]. We expect to tag at least 25 instances of each verb. If there are not enough occurrences in the Brown Corpus, we will use examples from the same sources as our extended corpus (see above).

## 5 Acknowledgements

Design and preparation of COMLEX Syntax has been supported by the Advanced Research Projects Agency through the Office of Naval Research under Awards No. MDA972-92-J-1016 and N00014-90-J-1851, and The Trustees of the University of Pennsylvania.